\documentclass[letterpaper]{article} 
\usepackage{aaai2026}  
\usepackage{times}  
\usepackage{helvet}  
\usepackage{courier}  
\usepackage[hyphens]{url}  
\usepackage{graphicx} 
\usepackage{natbib}  
\usepackage{caption} 
\usepackage{algorithm}
\usepackage{algorithmic}
\usepackage{tikz}
\usetikzlibrary{arrows.meta, calc}
\tikzset{>=latex}  
\usepackage{newfloat}
\usepackage{listings}
\DeclareCaptionStyle{ruled}{labelfont=normalfont,labelsep=colon,strut=off} 
\floatstyle{ruled}
\newfloat{listing}{tb}{lst}{}
\floatname{listing}{Listing}
 \nocopyright 

\title{Beyond Post-hoc Explanation: Toward Glassbox AI via Probabilistic Mediation}
\author {
    Manuele Leonelli
}
\affiliations{
    School of Science and Technology, IE University, Madrid, Spain\\
    manuele.leonelli@ie.edu
}

\begin{document}

\maketitle

\begin{abstract}
Large language models are rapidly becoming infrastructural components in high-stakes institutional settings, including public administration, legal reasoning, and healthcare, where opacity is not merely inconvenient but institutionally and legally untenable. Existing approaches to explainability are predominantly post-hoc, offering unstable, non-contestable accounts that have no formal relationship to the reasoning process that produced the output. We argue that the problem is not the absence of explanation but the absence of structured reasoning in the first place. This paper makes the case for a fundamentally different architecture, which we call the Glassbox Framework, in which Bayesian networks serve as transparent, ante-hoc mediation layers for generative models. Bayesian networks encode domain knowledge, causal assumptions, and probabilistic dependencies before inference occurs, enabling auditable reasoning traces, uncertainty quantification, and contestable outputs. We characterise the architecture of this framework and ground it in a benefit eligibility scenario, identifying the foundational challenges spanning semantic alignment, dynamic model construction, probabilistic grounding, and human governance that must be solved to realise it at scale. By shifting from post-hoc explanation to ante-hoc probabilistic mediation, this work outlines a principled path toward AI systems that are not only powerful but fundamentally accountable.
\end{abstract}


\section{Introduction}
Large language models (LLMs) have rapidly transitioned from research artefacts to consequential infrastructure, embedded in systems that make or inform decisions across public administration, healthcare, and legal reasoning \citep{bommasani2021opportunities}. This transition brings with it a governance challenge of the first order. When a generative model informs a bail decision, a benefit assessment, or a medical triage, the opacity of its reasoning is not merely a technical inconvenience, it is an institutional and legal problem. The EU Artificial Intelligence Act \citep{act2024eu} identifies a growing class of high-risk applications for which auditability and transparency are not optional, yet the dominant response to this challenge remains inadequate.

The governance stakes are not merely institutional abstractions. Public concern about AI-driven automation is widespread, growing, and unevenly distributed. Survey evidence from Germany shows that public support for AI regulation is far from settled, with systematic variation in whether current EU-level oversight is considered adequate \citep{cremaschi2025understanding}. Fear of job displacement due to AI is similarly prevalent across Latin America, structured by education, political ideology, and institutional trust \citep{cremaschi2025will}, and comparable anxieties have been documented across twenty countries and multiple domains of application \citep{dong2024fears}, echoing long-standing evidence that automation disproportionately threatens lower-skilled workers \citep{frey2017future}. These findings underscore that the opacity problem is not merely technical. When consequential decisions are made by systems whose reasoning cannot be examined or challenged, the populations most exposed to their effects are also those least equipped to contest them.

The dominant response to this challenge has been post-hoc explainability: the construction of a secondary model or procedure to approximate the behaviour of an opaque primary model after the fact \citep{ribeiro2016should, lundberg2017unified}. \citet{rudin2019stop} has argued forcefully that this approach is fundamentally flawed: explanations are unstable, often unfaithful to the underlying model, and provide no formal accountability guarantees. Her prescription, however, was written in a world where the choice between an interpretable model and a black box remained open. The emergence of LLMs as infrastructural components has closed that choice in many domains. One cannot replace an LLM with a decision tree. The black box is now the environment, not a modelling decision \citep{burrell2016machine}, and a new framework is needed.

We argue that the problem is not the absence of explanation but the absence of structured reasoning in the first place. This paper makes the case for a fundamentally different architecture, which we call the \textbf{Glassbox Framework}, in which Bayesian networks (BNs) \citep{pearl2009causality} serve as transparent, ante-hoc mediation layers for generative models. Rather than appending an explanation to an opaque output, the Glassbox Framework imposes a formally specified, inspectable reasoning structure before inference occurs, enabling outputs that are auditable, contestable, and grounded in explicit domain knowledge.

The paper makes four contributions. First, we introduce and formally characterise the Glassbox Framework as a conceptual architecture for accountable generative AI. Second, we define the BN-LLM interface as a scientific object in its own right, distinct from both post-hoc explainability and neuro-symbolic integration. Third, we characterise the representational gap between language and probability, illustrating the severity of the semantic alignment problem through a structured thought experiment grounded in normative domain reasoning, and demonstrate the framework's operational properties through a benefit eligibility scenario. Fourth, we identify the foundational research challenges that must be solved to realise the framework at scale, constituting a research agenda for probabilistic mediation in high-stakes AI systems.

\section{The Limits of Post-Hoc Explainability}

The dominant paradigm for addressing opacity in machine learning systems is post-hoc explainability: given a trained model and an input, a secondary procedure is applied to approximate or summarise the model's behaviour in human-interpretable terms \citep{ribeiro2016should, doshi2017towards,lundberg2017unified}. This paradigm has generated substantial research activity, but it rests on a flawed premise. The very notion of interpretability underlying these methods has been shown to be deeply underspecified: different stakeholders require different properties from an explanation, and no single post-hoc method satisfies them all \citep{lipton2016mythos}. The explanations produced are not accounts of how a system reasons, but approximations of what a system outputs, constructed after the fact by a separate mechanism with no formal connection to the original model. We identify three specific failure modes that are particularly consequential in high-stakes institutional settings.

\subsection{Instability}
Post-hoc explanations are sensitive to small perturbations in the input. Marginally different inputs can produce dramatically different explanations even when the underlying prediction is unchanged \citep{alvarez2018towards}. This instability is not a correctable limitation of specific methods; it is a structural consequence of approximating a complex function with a simpler surrogate. In institutional settings where consistent, reproducible reasoning is a legal and ethical requirement, an explanation that varies arbitrarily across similar cases is worse than no explanation at all, since it creates a false impression of transparency while providing none of its substance.

\subsection{Non-contestability}
A decision is contestable when an affected party can examine the reasoning that produced it, identify a specific point of disagreement, and mount a challenge on that basis \citep{wachter2017counterfactual, novelli2024accountability}. Post-hoc explanations cannot support genuine contestability because they have no formal relationship to the reasoning process that produced the output. Challenging a LIME or SHAP explanation is not challenging the model; it is challenging an approximation of the model, constructed by a different system, using different logic. Research in the social sciences establishes that explanations serve fundamentally social and dialogic functions as tools for managing accountability in human interactions, and that computational approximations of model behaviour do not satisfy the criteria that make explanations genuinely useful in institutional settings \citep{miller2019explanation}. Legal rights to explanation, as typically framed in data protection regulations, provide weak guarantees precisely because they do not require that explanations be faithful to the underlying model \citep{edwards2017slave}. The EU AI Act's requirements for transparency and human oversight in high-risk systems \citep{act2024eu} implicitly demand contestability of this kind, yet post-hoc methods are structurally incapable of providing it.

\subsection{The Accountability Gap}
Post-hoc methods tell you which input features influenced a prediction. They do not tell you whether that prediction is consistent with domain knowledge, whether it respects normative constraints, or whether the reasoning that produced it is defensible relative to any explicit standard \citep{rudin2019stop, buttaboni2026regulatory}. There is no explicit locus of responsibility, no structure that can be examined, challenged, or held to account. \citet{floridi2022unified} identify explicability as a foundational principle for trustworthy AI, but explicability requires more than a post-hoc summary. It requires that the reasoning itself be structured and inspectable.

These three failure modes share a common root: post-hoc explanation is a correction applied to a system that was never designed for accountability. The correction cannot substitute for the thing it was meant to replace. What is needed is not a better explanation but a fundamentally different architecture: one in which structured, inspectable reasoning is built into the system before inference occurs, not appended to it afterwards.

\section{The Glassbox Framework}
The limitations previously identified motivate a shift in how accountability is conceived in AI systems. Rather than asking how to explain a decision after it has been made, we ask how to ensure that the reasoning process itself is structured, inspectable, and formally grounded before inference occurs. We call this shift \textbf{ante-hoc accountability}, and the architecture that realises it the \textbf{Glassbox Framework}.

The distinction between post-hoc and ante-hoc accountability is not merely terminological. Post-hoc approaches treat the reasoning process as fixed and opaque, and attempt to reconstruct it after the fact using a secondary system. Ante-hoc approaches treat the reasoning structure as a first-class object, specified explicitly before inference begins. The accountability properties of the system (transparency, contestability, uncertainty quantification) are then direct consequences of the architecture, not approximations derived from it. This is the sense in which the Glassbox Framework differs fundamentally from explainable AI: it does not explain opacity, it replaces it.

Retrieval-augmented generation \citep{lewis2020retrieval} represents one attempt to ground LLM outputs in external knowledge, but retrieved content remains subject to the LLM's unstructured reasoning: there is no formal inference, no consistency enforcement, and no inspectable reasoning trace. The Glassbox Framework differs fundamentally in that the BN, not the LLM, is the reasoning authority.

\subsection{Conceptual Foundations}

The Glassbox Framework rests on three conceptual commitments. The first is that structured probabilistic knowledge, not natural language fluency, should be the substrate of consequential reasoning. LLMs are powerful generators of plausible text, but plausibility is not the same as correctness, consistency, or accountability \citep{bender2021dangers, weidinger2021ethical}. When a system informs a high-stakes decision, the reasoning that produced that decision must be traceable to explicit assumptions, not implicit statistical patterns learned from web-scale data.

The second commitment is that BNs are the natural formal candidate for this role \citep{pearl2009causality, fenton2018risk}. A BN encodes a joint probability distribution over a set of variables as a directed acyclic graph, making conditional dependencies explicit and inspectable. This structure supports three properties essential for ante-hoc accountability. \textbf{Uncertainty quantification} is native: every inference produces a probability distribution, not a point prediction, and the contribution of each variable to the final output can be traced and examined \citep{koller2009probabilistic, ballester2022computing}. \textbf{Counterfactual reasoning} is direct: the effect of intervening on any variable, or removing any piece of evidence, can be computed from the graph structure without requiring a secondary approximation \citep{peters2017elements}. \textbf{Modularity} is preserved: domain-specific normative knowledge can be encoded in the BN without affecting the general architecture, making the framework portable across institutional settings \citep{neil2000building, leonelli2020coherent}.

The third commitment is that the interface between language and probability is a scientific object in its own right, not an engineering detail. The BN-LLM interface is the point at which two fundamentally different representations of knowledge must be made to interact coherently: the distributional, implicit, high-dimensional representations of a generative model, and the discrete, structured, semantically grounded representations of a probabilistic graphical model. How this interface is constructed, validated, and governed is, we argue, the central open problem in accountable generative AI, and one that we examine in depth later in this paper.

\subsection{Architecture}

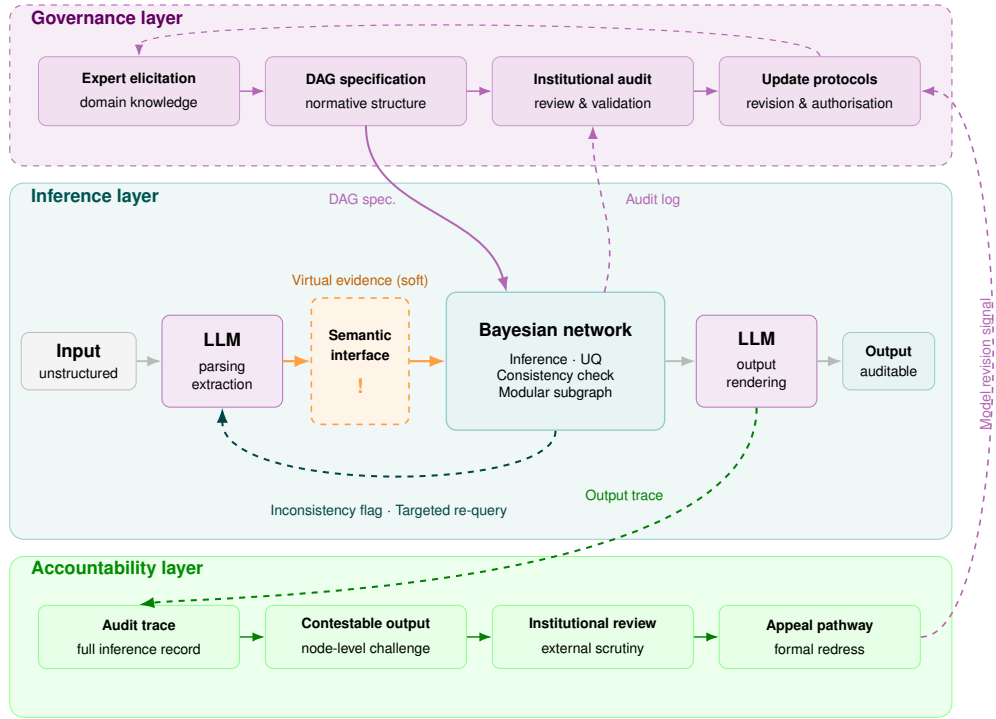
\begin{figure*}
    \centering
\scalebox{0.76}{
\begin{tikzpicture}[font=\sffamily, >=Latex]

\colorlet{govfill}{violet!7}
\colorlet{govbord}{violet!55}
\colorlet{govtxt}{violet!75!black}
\colorlet{govnode}{violet!12}
\colorlet{govnbord}{violet!40}
\colorlet{inffill}{teal!6}
\colorlet{infbord}{teal!40}
\colorlet{inftxt}{teal!65!black}
\colorlet{accfill}{green!7}
\colorlet{accbord}{green!45}
\colorlet{acctxt}{green!55!black}
\colorlet{accnode}{green!10}
\colorlet{accnbord}{green!45}
\colorlet{sembord}{orange!75}
\colorlet{semfill}{orange!9}
\colorlet{llmfill}{violet!9}
\colorlet{bnfill}{teal!9}
\colorlet{gfill}{gray!9}
\colorlet{purparr}{violet!60}
\colorlet{grnarr}{green!50!black}
\colorlet{feedarr}{teal!55!black}

\fill[govfill, rounded corners=7pt]
  (0.3, 9.9) rectangle (16.7, 12.7);
\draw[govbord, dashed, line width=0.6pt, rounded corners=7pt]
  (0.3, 9.9) rectangle (16.7, 12.7);

\fill[inffill, rounded corners=7pt]
  (0.3, 3.4) rectangle (16.7, 9.6);
\draw[infbord, line width=0.4pt, rounded corners=7pt]
  (0.3, 3.4) rectangle (16.7, 9.6);

\fill[accfill, rounded corners=7pt]
  (0.3, 0.3) rectangle (16.7, 3.1);
\draw[accbord, line width=0.4pt, rounded corners=7pt]
  (0.3, 0.3) rectangle (16.7, 3.1);

\node[anchor=west, font=\sffamily\small\bfseries, text=govtxt]
  at (0.55, 12.45) {Governance layer};
\node[anchor=west, font=\sffamily\small\bfseries, text=inftxt]
  at (0.55, 9.35) {Inference layer};
\node[anchor=west, font=\sffamily\small\bfseries, text=acctxt]
  at (0.55, 2.88) {Accountability layer};

\node[draw=govnbord, fill=govnode, rounded corners=4pt, line width=0.5pt,
  minimum width=3.5cm, minimum height=1.2cm, align=center]
  (gelicit) at (2.55, 11.2) {
    {\sffamily\scriptsize\bfseries Expert elicitation}\\[1pt]
    {\sffamily\scriptsize domain knowledge}};

\node[draw=govnbord, fill=govnode, rounded corners=4pt, line width=0.5pt,
  minimum width=3.5cm, minimum height=1.2cm, align=center]
  (gdag) at (6.50, 11.2) {
    {\sffamily\scriptsize\bfseries DAG specification}\\[1pt]
    {\sffamily\scriptsize normative structure}};

\node[draw=govnbord, fill=govnode, rounded corners=4pt, line width=0.5pt,
  minimum width=3.5cm, minimum height=1.2cm, align=center]
  (gaudit) at (10.45, 11.2) {
    {\sffamily\scriptsize\bfseries Institutional audit}\\[1pt]
    {\sffamily\scriptsize review \& validation}};

\node[draw=govnbord, fill=govnode, rounded corners=4pt, line width=0.5pt,
  minimum width=3.5cm, minimum height=1.2cm, align=center]
  (gupdate) at (14.40, 11.2) {
    {\sffamily\scriptsize\bfseries Update protocols}\\[1pt]
    {\sffamily\scriptsize revision \& authorisation}};

\node[draw=accnbord, fill=accnode, rounded corners=4pt, line width=0.5pt,
  minimum width=3.5cm, minimum height=1.1cm, align=center]
  (aaudit) at (2.55, 1.7) {
    {\sffamily\scriptsize\bfseries Audit trace}\\[1pt]
    {\sffamily\scriptsize full inference record}};

\node[draw=accnbord, fill=accnode, rounded corners=4pt, line width=0.5pt,
  minimum width=3.5cm, minimum height=1.1cm, align=center]
  (acontest) at (6.50, 1.7) {
    {\sffamily\scriptsize\bfseries Contestable output}\\[1pt]
    {\sffamily\scriptsize node-level challenge}};

\node[draw=accnbord, fill=accnode, rounded corners=4pt, line width=0.5pt,
  minimum width=3.5cm, minimum height=1.1cm, align=center]
  (areview) at (10.45, 1.7) {
    {\sffamily\scriptsize\bfseries Institutional review}\\[1pt]
    {\sffamily\scriptsize external scrutiny}};

\node[draw=accnbord, fill=accnode, rounded corners=4pt, line width=0.5pt,
  minimum width=3.5cm, minimum height=1.1cm, align=center]
  (aappeal) at (14.40, 1.7) {
    {\sffamily\scriptsize\bfseries Appeal pathway}\\[1pt]
    {\sffamily\scriptsize formal redress}};

\node[draw=gray!50, fill=gfill, rounded corners=4pt, line width=0.5pt,
  minimum width=2.0cm, minimum height=1.0cm, align=center]
  (input) at (1.5, 6.5) {
    \sffamily\small\bfseries Input\\[-1pt]
    \sffamily\scriptsize unstructured};

\node[draw=govbord, fill=llmfill, rounded corners=4pt, line width=0.5pt,
  minimum width=2.1cm, minimum height=1.6cm, align=center]
  (llmparse) at (4.0, 6.5) {
    \sffamily\small\bfseries LLM\\[1pt]
    \sffamily\scriptsize parsing\\[-2pt]
    \sffamily\scriptsize extraction};

\node[draw=sembord, dashed, fill=semfill, rounded corners=4pt, line width=0.8pt,
  minimum width=1.7cm, minimum height=2.2cm, align=center]
  (semantic) at (6.4, 6.5) {
    \sffamily\scriptsize\bfseries Semantic\\[-1pt]
    \sffamily\scriptsize\bfseries interface\\[5pt]
    {\color{sembord}\normalsize\bfseries !}};

\node[draw=infbord, fill=bnfill, rounded corners=5pt, line width=0.7pt,
  minimum width=3.8cm, minimum height=2.4cm, align=center]
  (bn) at (9.8, 6.5) {
    \sffamily\small\bfseries Bayesian network\\[3pt]
    \sffamily\scriptsize Inference $\cdot$ UQ\\[-3pt]
    \sffamily\scriptsize Consistency check\\[-3pt]
    \sffamily\scriptsize Modular subgraph};

\node[draw=govbord, fill=llmfill, rounded corners=4pt, line width=0.5pt,
  minimum width=2.1cm, minimum height=1.6cm, align=center]
  (llmout) at (13.3, 6.5) {
    \sffamily\small\bfseries LLM\\[1pt]
    \sffamily\scriptsize output\\[-2pt]
    \sffamily\scriptsize rendering};

\node[draw=infbord, fill=bnfill, rounded corners=4pt, line width=0.5pt,
  minimum width=1.6cm, minimum height=1.0cm, align=center]
  (audout) at (15.6, 6.5) {
    \sffamily\scriptsize\bfseries Output\\[-1pt]
    \sffamily\scriptsize auditable};

\draw[->, gray!55, line width=0.9pt]
  (input.east) -- (llmparse.west);
\draw[->, sembord, line width=1.1pt]
  (llmparse.east) -- (semantic.west);
\draw[->, sembord, line width=1.1pt]
  (semantic.east) -- (bn.west);
\draw[->, gray!55, line width=0.9pt]
  (bn.east) -- (llmout.west);
\draw[->, gray!55, line width=0.9pt]
  (llmout.east) -- (audout.west);

\draw[->, purparr, line width=0.7pt]
  (gelicit.east) -- (gdag.west);
\draw[->, purparr, line width=0.7pt]
  (gdag.east) -- (gaudit.west);
\draw[->, purparr, line width=0.7pt]
  (gaudit.east) -- (gupdate.west);

\draw[->, dashed, purparr, line width=0.6pt]
  (gupdate.north)
  .. controls (14.40, 12.5) and (2.55, 12.5) ..
  (gelicit.north);

\draw[->, grnarr, line width=0.7pt]
  (aaudit.east) -- (acontest.west);
\draw[->, grnarr, line width=0.7pt]
  (acontest.east) -- (areview.west);
\draw[->, grnarr, line width=0.7pt]
  (areview.east) -- (aappeal.west);

\draw[->, dashed, feedarr, line width=1.0pt]
  (bn.south)
  .. controls (9.8, 4.2) and (4.0, 4.2) ..
  (llmparse.south);
\node[font=\sffamily\scriptsize, text=feedarr, anchor=center]
  at (6.9, 3.88)
  {Inconsistency flag $\cdot$ Targeted re-query};

\node[font=\sffamily\scriptsize, text=orange!75!black, anchor=south]
  at (6.4, 7.65) {Virtual evidence (soft)};

\draw[->, purparr, line width=1.0pt]
  (6.50, 10.60) to[out=270, in=105] (8.95, 7.70);
\node[font=\sffamily\scriptsize, text=purparr, anchor=east]
  at (7.15, 9.30) {DAG spec.};

\draw[->, dashed, purparr, line width=0.8pt]
  (10.65, 7.70) to[out=75, in=270] (10.45, 10.60);
\node[font=\sffamily\scriptsize, text=purparr, anchor=west]
  at (10.90, 9.30) {Audit log};

\draw[->, dashed, grnarr, line width=1.0pt]
  (llmout.south)
  .. controls (13.3, 2.4) and (3.5, 2.4) ..
  (aaudit.north);
\node[font=\sffamily\scriptsize, text=grnarr, anchor=center]
  at (11.0, 4.15) {Output trace};
\draw[->, dashed, purparr, line width=0.8pt]
  (aappeal.east)
  .. controls (17.8, 1.7) and (17.8, 11.2) ..
  (gupdate.east);
\node[font=\sffamily\scriptsize, text=purparr, rotate=90, anchor=south]
  at (17.55, 6.45) {Model revision signal};

\end{tikzpicture}}
\caption{The Glassbox Framework. The \textbf{governance layer} (top) cycles through expert elicitation, DAG specification, audit, and revision. The \textbf{inference layer} (middle) mediates LLM-BN interaction via the semantic translation interface, with virtual soft evidence entering the BN and inconsistency flags routed back for re-query. The \textbf{accountability layer} (bottom) sequences the output trace through audit, contestation, review, and appeal, with a model revision signal closing the institutional loop.}
    \label{fig:architecture}
\end{figure*}

The Glassbox Framework is organized into three vertically arranged functional layers, each with distinct responsibilities and accountability implications. Figure~\ref{fig:architecture} illustrates the overall structure.

\subsubsection{The Governance Layer.}
The governance layer is institutional rather than computational. Domain experts, legal authorities, or institutional auditors specify the BN that anchors the inference layer: which variables are included, what causal structure is encoded in the DAG, and what prior distributions reflect current institutional standards. This layer also receives audit logs from the inference layer below, making the system's reasoning history available for external scrutiny and challenge. Who governs this layer, under what authority, and through what protocols are not engineering questions. They are institutional ones whose answers matter as much as any technical design decision. Where observational data are available, structure learning algorithms may complement expert elicitation, though the defensibility of data-driven structures raises questions we examine later.

Governance does not end at specification. The layer also establishes the update protocol: the conditions under which the BN may be revised, who authorises revision, and how changes are communicated to affected parties. This makes governance not a one-time act but an ongoing institutional process.

\subsubsection{The Inference Layer.} The inference layer is the technical core of the framework and operates through an iterative horizontal interaction between an LLM and a BN, mediated by a semantically uncertain interface. An LLM first receives unstructured input and performs an initial parsing and extraction pass, proposing candidate assignments for the variables defined in the BN. These proposals pass through the \textbf{semantic translation interface}, where natural language representations must be mapped onto the discrete, formally specified variable space of the BN. This interface is not a solved component: it is the most technically uncertain element of the architecture, and we mark it explicitly as such in Figure~\ref{fig:architecture}.

At the centre of the inference layer sits the BN itself, functioning as a dynamic, modular reasoning object. It is not invoked in its entirety for every query. Rather, the contextually relevant subgraph is activated and the remaining variables are marginalized, focusing computational effort and normative structure on what matters for the decision at hand. Critically, subgraph selection cannot itself depend on the LLM's initial parsing, since a parsing failure would marginalize out precisely the variables needed to detect it. The framework therefore defaults to a normatively dense base activation (in benefit eligibility reasoning, for instance, nodes encoding employment status, income, residency, and contribution record are always included) and pruning occurs only once the BN has confirmed that the excluded variables are conditionally independent of the active subgraph given the current evidence. This conservative initialization breaks the bootstrap dependency between parsing and subgraph selection.

The BN performs probabilistic inference over the instantiated subgraph \citep{koller2009probabilistic}, quantifies uncertainty at each node \citep{ballester2022computing}, and checks the consistency of the LLM-supplied assignments against the encoded domain structure \citep{fenton2018risk}. When assignments are incoherent, the BN generates a targeted re-query, routed back through the semantic translation interface to the LLM for focused re-examination. This feedback loop is iterative: it runs until a coherent instantiation is achieved, or until an irresolvable conflict signals a structurally informative failure. The posterior distribution and full inference trace are then passed to an output rendering LLM, which translates the structured result back into natural language. The result is not a bare prediction but an auditable account of the reasoning process. Beyond this iterative re-querying, the interaction between LLM and BN may in principle extend to model revision: flagged inconsistencies may signal that the current DAG requires a new variable, or that existing conditional probabilities require updating in light of evidence not anticipated at specification time. Such revisions engage the governance layer directly, since any structural change to the BN is a normative decision requiring institutional authorisation rather than a purely technical update.

\subsubsection{The Accountability Layer.} The accountability layer receives this output and makes the full reasoning trace available for institutional use: appeal, review, or challenge. Because the trace is a direct record of the ante-hoc structure rather than a post-hoc approximation, contestability is meaningful: an affected party can identify a specific node, conditional dependency, or governance decision as the basis of a challenge \citep{novelli2024accountability}. Algorithmic accountability requires not just transparency but procedural legitimacy: the ability to examine, contest, and correct reasoning through institutional channels. This requires that the reasoning process itself be formally specified before deployment \citep{kroll2017accountable}. The accountability layer also serves a prospective function: accumulated audit logs make patterns of inconsistency visible over time, enabling institutional review of whether the BN's encoded assumptions remain adequate as the deployment context evolves.

Three hard problems remain visible without resolution. Subgraph selection cannot fully precede parsing without circularity. Translating LLM outputs into BN likelihoods requires bridging incompatible probabilistic regimes. And the termination conditions for the feedback loop require formal characterisation. We examine each in the following sections.

\subsection{Key Properties}

The Glassbox Framework possesses four properties that arise directly from its architecture rather than from any secondary approximation system.

\subsubsection{Transparency of Structure.} In the Glassbox Framework, the reasoning structure is fully specified before inference occurs. The variables, their conditional dependencies, and the prior distributions that govern inference are encoded in the BN and available for inspection at any point. There is no secondary model approximating the behaviour of a primary one. The structure that produces the output is the structure that can be examined, and these are the same object \citep{doshi2017towards,rudin2019stop}. Where the LLM-BN interaction reveals gaps in the current structure (e.g. variables not anticipated at specification time, or dependencies not encoded) revision is possible but only through the governance layer, preserving the auditability of the structure at every stage.

\subsubsection{Contestability of Outputs.} Because the reasoning trace is a direct record of the inference process rather than a post-hoc reconstruction, it supports genuine contestability \citep{wachter2017counterfactual, novelli2024accountability}. An affected party, an auditor, or a legal authority can identify a specific node, a specific conditional dependency, or a specific governance decision encoded in the DAG as the basis of a challenge. This is not possible with LIME or SHAP explanations, which have no formal relationship to the model they purport to explain.

\subsubsection{Formal Uncertainty Quantification.} Every inference in a BN produces a probability distribution over outcome variables, not a point prediction \citep{koller2009probabilistic}. The contribution of each variable to the posterior can be quantified through sensitivity analysis, making the sources of uncertainty explicit and traceable \citep{ballester2025global}. In high-stakes settings where decisions must be justified under uncertainty, this property is not a convenience: it is a governance requirement.

\subsubsection{Modularity Across Domains.} The BN encoding an administrative eligibility domain is not the same as the BN encoding a medical or judicial domain, and domain-specific normative knowledge can be incorporated without altering the general architecture \citep{neil2000building}. This modularity means that the Glassbox Framework is not a domain-specific solution but a general architecture instantiated through domain-specific knowledge. It also means that governance responsibilities are clearly localised: the domain BN is the object that domain experts specify, validate, and update, independently of the LLM components that surround it. Modularity does not imply rigidity. The BN structure is fixed at any given moment of deployment, this is what makes it inspectable and contestable, but it is revisable through governed processes when the domain evolves, new evidence types emerge, or operational experience reveals gaps. The key distinction is between revision, which is an institutional act requiring authorisation and documentation, and inference, which proceeds over a stable, auditable structure. Modularity ensures that such revisions remain domain-contained: a change to the eligibility BN does not propagate to the medical or judicial instantiation of the framework.

Table~\ref{tab:comparison} synthesises the structural differences between post-hoc explainability and the Glassbox Framework across the dimensions most consequential for high-stakes institutional deployment. The contrast is not one of degree but of kind: the properties of the Glassbox Framework are direct consequences of its architecture, not approximations derived from a secondary system.

\begin{table}
\centering
\caption{Structural comparison between post-hoc explainability
         and the Glassbox Framework across dimensions relevant
         to high-stakes institutional deployment.}
\label{tab:comparison}
\scalebox{0.82}{
\begin{tabular}{lcc}
\hline
\textbf{Property}
  & \textbf{Post-hoc XAI}
  & \textbf{Glassbox} \\
\hline
Reasoning specified before inference
  & No & Yes \\
Faithful to underlying reasoning
  & Not guaranteed & Structural \\
Stable across similar inputs
  & No & Yes \\
Supports node-level contestation
  & No & Yes \\
Formal uncertainty quantification
  & Approximate & Native \\
Counterfactual reasoning
  & Approximate & Direct \\
Full inference trace
  & Absent & Auditable \\
Governance layer
  & External & Embedded \\
Institutional updatability
  & Ad hoc & Governed \\
\hline
\end{tabular}}
\end{table}

\section{The Representational Gap: Where Language Meets Probability}

\label{sec:gap}

The Glassbox Framework is not a system that can be built by assembling existing components. The architecture previously described identifies three points of genuine difficulty: the semantic translation interface, the construction and maintenance of the BN, and the probabilistic grounding of LLM outputs. Each of these corresponds to a research problem that is currently open, and each is harder than it appears. We characterise them here not as engineering challenges but as foundational problems whose resolution requires new conceptual and mathematical machinery.

\subsection{Semantic Alignment}
\begin{figure*}
    \centering

\begin{tikzpicture}[font=\sffamily, >=Latex]

\colorlet{langfill}{blue!5}
\colorlet{langbord}{blue!35}
\colorlet{intfill}{orange!8}
\colorlet{intbord}{orange!65}
\colorlet{bnfill}{teal!5}
\colorlet{bnbord}{teal!35}
\colorlet{clrfill}{teal!8}
\colorlet{clrbord}{teal!50}
\colorlet{ambfill}{orange!10}
\colorlet{ambbord}{orange!60}
\colorlet{unmfill}{red!8}
\colorlet{unmbord}{red!50}
\colorlet{clrarr}{teal!65!black}
\colorlet{ambarr}{orange!80!black}
\colorlet{unmarr}{red!55!black}


\fill[langfill, rounded corners=5pt] (0,0) rectangle (3.6,8);
\draw[langbord, line width=0.4pt, rounded corners=5pt]
  (0,0) rectangle (3.6,8);

\fill[intfill, rounded corners=5pt] (3.7,0) rectangle (6.1,8);
\draw[intbord, dashed, line width=0.7pt, rounded corners=5pt]
  (3.7,0) rectangle (6.1,8);

\fill[bnfill, rounded corners=5pt] (6.2,0) rectangle (14.0,8);
\draw[bnbord, line width=0.4pt, rounded corners=5pt]
  (6.2,0) rectangle (14.0,8);


\node[anchor=west, font=\sffamily\small\bfseries, text=blue!65!black]
  at (0.20, 7.72) {Language space};
\node[anchor=west, font=\sffamily\tiny, text=blue!55!black]
  at (0.20, 7.44) {Claim document expressions};

\node[font=\sffamily\small\bfseries, text=orange!75!black]
  at (4.90, 7.72) {Semantic};
\node[font=\sffamily\small\bfseries, text=orange!75!black]
  at (4.90, 7.44) {interface};

\node[anchor=west, font=\sffamily\small\bfseries, text=teal!65!black]
  at (6.40, 7.72) {BN variable space};
\node[anchor=west, font=\sffamily\tiny, text=teal!55!black]
  at (6.40, 7.44) {Formally specified nodes};


\node[
  draw=clrbord, fill=clrfill, rounded corners=3pt, line width=0.6pt,
  minimum width=3.20cm, minimum height=0.60cm,
  font=\sffamily\scriptsize\itshape
] (E1) at (1.80, 6.28) {"currently unemployed"};

\node[
  draw=clrbord, fill=clrfill, rounded corners=3pt, line width=0.6pt,
  minimum width=3.20cm, minimum height=0.60cm,
  font=\sffamily\scriptsize\itshape
] (E2) at (1.80, 5.43) {"no active employment"};

\node[
  draw=clrbord, fill=clrfill, rounded corners=3pt, line width=0.6pt,
  minimum width=3.20cm, minimum height=0.60cm,
  font=\sffamily\scriptsize\itshape
] (E3) at (1.80, 4.59) {"jobless since March"};

\node[
  draw=ambbord, fill=ambfill, rounded corners=3pt, line width=0.6pt,
  dashed, minimum width=3.20cm, minimum height=0.60cm,
  font=\sffamily\scriptsize\itshape
] (E4) at (1.80, 3.56) {"freelance occasionally"};

\node[
  draw=ambbord, fill=ambfill, rounded corners=3pt, line width=0.6pt,
  dashed, minimum width=3.20cm, minimum height=0.60cm,
  font=\sffamily\scriptsize\itshape
] (E5) at (1.80, 2.59) {"no salary received"};

\node[
  draw=unmbord, fill=unmfill, rounded corners=3pt, line width=0.6pt,
  dashed, minimum width=3.20cm, minimum height=0.60cm,
  font=\sffamily\scriptsize\itshape
] (E6) at (1.80, 1.54) {"between opportunities"};


\fill[clrfill] (0.25,0.52) rectangle (0.50,0.72);
\draw[clrbord, line width=0.5pt] (0.25,0.52) rectangle (0.50,0.72);
\node[anchor=west, font=\sffamily\tiny, text=teal!60!black]
  at (0.56, 0.62) {Clear mapping};

\fill[ambfill] (0.25,0.22) rectangle (0.50,0.42);
\draw[ambbord, dashed, line width=0.5pt] (0.25,0.22) rectangle (0.50,0.42);
\node[anchor=west, font=\sffamily\tiny, text=orange!70!black]
  at (0.56, 0.32) {Uncertain / ambiguous};


\node[
  draw=teal!50, fill=teal!9, rounded corners=4pt, line width=0.6pt,
  minimum width=7.20cm, minimum height=1.20cm, align=center
] (V1) at (10.10, 5.96) {
  \sffamily\small\bfseries EmploymentStatus\\[2pt]
  \sffamily\tiny employed $\cdot$ unemployed $\cdot$ inactive
};

\node[
  draw=teal!50, fill=teal!9, rounded corners=4pt, line width=0.6pt,
  minimum width=7.20cm, minimum height=1.20cm, align=center
] (V2) at (10.10, 2.77) {
  \sffamily\small\bfseries IncomeReceived\\[2pt]
  \sffamily\tiny received $\cdot$ not received
};


\node[font=\sffamily\tiny\itshape, text=teal!60!black]
  at (10.10, 5.10) {Many expressions, one variable};

\node[font=\sffamily\tiny\itshape, text=orange!65!black]
  at (10.10, 4.22) {One expression, two candidate nodes};

\node[font=\sffamily\tiny\itshape, text=orange!55!black]
  at (10.10, 1.92) {Indirect signal, uncertain polarity};


\draw[intbord, line width=1.0pt]
  (4.90,0.82) -- (4.48,0.18) -- (5.32,0.18) -- cycle;
\node[font=\sffamily\bfseries\scriptsize, text=orange!80!black]
  at (4.90, 0.40) {!};
\node[font=\sffamily\tiny, text=orange!65!black]
  at (4.90, 1.00) {Unresolved};


\draw[->, clrarr, line width=1.0pt]
  (3.40, 6.28) -- (6.50, 6.10);

\draw[->, clrarr, line width=1.0pt]
  (3.40, 5.43) -- (6.50, 5.96);

\draw[->, clrarr, line width=1.0pt]
  (3.40, 4.59) -- (6.50, 5.80);

\draw[ambarr, line width=1.0pt, dashed]
  (3.40, 3.56) -- (4.85, 3.56);

\filldraw[ambarr] (4.85, 3.56) circle (2.2pt);

\draw[->, ambarr, line width=1.0pt, dashed]
  (4.85, 3.56) -- (6.50, 5.50);

\draw[ambarr, line width=1.0pt, dashed]
  (4.85, 3.56) -- (4.85, 2.92);
\draw[->, ambarr, line width=1.0pt, dashed]
  (4.85, 2.92) -- (6.50, 2.92);

\draw[->, ambarr, line width=1.0pt, dashed]
  (3.40, 2.59) -- (6.50, 2.77);

\draw[unmarr, line width=1.0pt, dashed]
  (3.40, 1.54) -- (4.72, 1.70);

\draw[unmarr, line width=1.8pt]
  (4.56, 1.54) -- (4.88, 1.86);
\draw[unmarr, line width=1.8pt]
  (4.88, 1.54) -- (4.56, 1.86);

\node[anchor=west, font=\sffamily\tiny\itshape, text=red!55!black]
  at (4.96, 1.90) {unmappable};

\end{tikzpicture}
 \caption{The semantic alignment problem. Natural language expressions from a benefit eligibility document must be mapped onto formally specified BN variables. Clear expressions (solid teal arrows) map unambiguously to a single variable, illustrating how multiple surface forms can converge on one node. An ambiguous expression (dashed amber, forking arrow) maps to two candidate nodes simultaneously, with no principled criterion for resolution. An indirect expression connects to a variable with uncertain polarity. One expression resists variable-level encoding entirely. This interface between language and probability is currently unsolved at scale and constitutes the central technical challenge of the Glassbox Framework.}
    \label{fig:semantic}
\end{figure*}
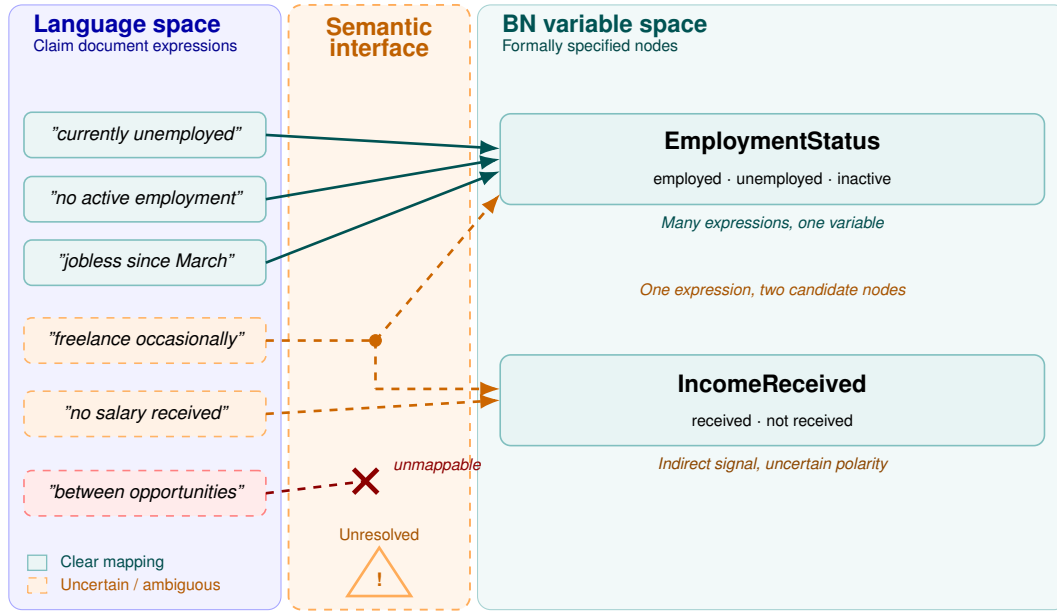

The semantic translation interface requires mapping natural language representations onto a formally specified, discrete variable space. This is not a retrieval problem. It is a problem of representational alignment between two knowledge systems that are fundamentally incommensurable: one that is continuous, distributional, and context-dependent, and one that is discrete, structured, and semantically grounded. Figure~\ref{fig:semantic} illustrates these mapping types in a benefit eligibility context.

The difficulty has three distinct dimensions. First, the same evidential concept can be expressed in radically different ways across documents, speakers, and domains. In benefit eligibility reasoning, for instance, the concept of employment status may surface as explicit declarations, as behavioural descriptions, as negated claims, or as inferences drawn from indirect financial evidence, each requiring a different extraction strategy but mapping to the same BN variable. Second, the reverse problem is equally serious: the same surface expression may correspond to different variables depending on context, making a naive lexical matching approach not merely imprecise but structurally wrong. Third, some concepts that are institutionally central may resist variable-level encoding entirely, either because they are inherently continuous or because their normative meaning cannot be captured by a binary or categorical indicator. As illustrated in Figure~\ref{fig:semantic}, the same concept may be expressed through multiple surface forms that must converge on a single node, a single expression may fork ambiguously toward competing nodes, or an expression may resist variable-level encoding entirely.

Consider a normatively structured domain where a key condition for a consequential outcome (e.g. a mental state required for criminal liability, or a contribution threshold required for benefit eligibility) must be inferred from unstructured narrative text. Even a carefully designed extraction pipeline may systematically fail to identify the relevant language, producing parameter estimates that invert the expected relationship: the condition estimated as more probable precisely when the textual signal for it is absent. This failure mode is not detectable from the network's structural properties alone. It emerges only at the interface between language and probability, and it has no solution within existing extraction or retrieval methods. The implication is sobering: a BN that is structurally sound and formally correct can produce systematically wrong inferences not because its reasoning structure is flawed, but because the language pipeline that feeds it cannot reliably identify the concepts the structure was designed to reason about.

\subsection{Dynamic Model Construction}

The BN in the Glassbox Framework is not a static object. In any realistic deployment, the domain it encodes will evolve: legal standards change, new evidence types emerge, institutional definitions are revised, and the normative structure that the DAG is meant to represent may shift over time. This raises questions that go well beyond model fitting.

Who constructs the initial DAG? Expert elicitation is the standard answer \citep{fenton2018risk}, but it is costly, slow, and subject to disagreement among experts whose normative commitments may differ. Structure learning from data is an alternative \citep{koller2009probabilistic}, but in normative domains it produces models that are empirically grounded without being normatively defensible: a model that learns the statistical regularities of past benefit denial decisions is not the same as a model that encodes the statutory criteria for benefit eligibility. The two may diverge in precisely the cases where the distinction matters most.

How is the DAG updated when the domain changes? Incremental structure learning under distributional shift is an active area of research, but it has not been applied to domains where the update itself is a policy decision rather than a statistical one. And how does the system detect when the current DAG is no longer adequate, when the distribution of incoming inputs has moved outside the region of the domain the BN was designed to represent? These are open problems that connect to structure learning \citep{koller2009probabilistic}, active learning, and model criticism literatures, but none of these literatures has addressed the normative dimension that the Glassbox Framework requires.

\subsection{Probabilistic Grounding}

LLMs do not produce calibrated probabilities in any sense that is compatible with the requirements of a BN. The softmax outputs of a language model are not posterior probabilities over the variables of interest. They are scores over a vocabulary, shaped by training objectives that have no formal relationship to the conditional probability tables the BN requires. Bridging these two probabilistic regimes is the mathematical core of the framework and the problem for which existing techniques are least adequate.

The difficulty is not simply one of calibration in the standard sense: adjusting a model's confidence scores to match empirical frequencies \citep{doshi2017towards}. It is a problem of translating between two fundamentally different representations of uncertainty. An LLM expresses uncertainty through the distribution of its token-level outputs; a BN expresses uncertainty through conditional probability distributions over explicitly defined variables. These are not two representations of the same thing. They are representations of different things, and the question of how to construct a principled mapping between them, one that preserves the formal properties required for coherent probabilistic inference, is not answered by any existing framework.

One partial specification of the interface is worth stating explicitly, as it constrains the design space. LLM extractions should not enter the BN as hard evidence: deterministic instantiations that set a node to a fixed state with probability one. Hard evidence would inherit the LLM's classification errors directly, bypassing the BN's structural priors and defeating the purpose of the mediation layer. Instead, LLM outputs should be treated as virtual evidence \citep{pearl2009causality}: likelihood ratios that shift the probability of a node's state without overriding the BN's prior structure. Under virtual evidence, an uncalibrated LLM confidence spike is damped by the domain knowledge encoded in the DAG, and the BN's structural priors can mathematically suppress assignments that are incoherent given the surrounding evidence. This framing preserves the BN's role as a reasoning authority rather than reducing it to a downstream filter on LLM outputs. The formal conditions under which virtual evidence can be derived from LLM token distributions, however, remain an open problem.

A related difficulty concerns the iterative feedback loop described in the architecture. For the loop to terminate coherently, there must be a well-defined criterion for when the LLM-supplied variable assignments are sufficiently consistent with the BN's encoded structure to proceed to inference. Defining this criterion formally, in a way that is neither so strict that the loop never terminates nor so permissive that incoherent assignments propagate into the reasoning trace, requires a theory of the interface that does not currently exist.

\subsection{Human Governance}

The governance layer of the Glassbox Framework is not a technical component. It is an institutional one, and the problems it raises are correspondingly different in character. Who specifies the DAG? Under what authority? By what process are disagreements among domain experts resolved? How is the BN validated before deployment, and by whom? How are updates to the DAG governed, audited, and communicated to affected parties?

These questions do not have technical answers, but they have structural ones. The EU AI Act \citep{act2024eu} establishes conformity assessment requirements for high-risk AI systems, but it does not specify what the governance of a probabilistic reasoning layer should look like in practice \citep{buttaboni2026regulatory}. The accountability literature identifies contestability as a requirement \citep{novelli2024accountability}, but it does not provide formal protocols for the kind of structured, node-level challenge that the Glassbox Framework makes possible. End-to-end auditing frameworks treat accountability as an organisational process rather than a model property, requiring formal documentation, internal review mechanisms, and external audit pathways \citep{raji2020closing}. The Glassbox Framework's governance layer is designed to be precisely this kind of auditable structure.

High-level ethical principles are insufficient without institutional mechanisms that give them operational force \citep{mittelstadt2019principles}. Moreover, encoding social and normative reasoning into formal structures risks what has been called the abstraction trap: losing the social context that gives decisions their meaning \citep{selbst2019fairness}. This risk is heightened in the Glassbox Framework precisely because the BN concentrates normative choices into an explicit structure, making the consequences of misconfiguration both more visible and more consequential.

Crucially, governance failure at this layer propagates silently through the rest of the system. A misconfigured DAG produces inference traces that are internally coherent but normatively wrong. Because the traces are structured and auditable, they may command more institutional authority than a bare LLM output, making the consequences of governance failure more serious, not less. The Glassbox Framework makes governance both more important and more tractable: more important because the BN concentrates normative choices into an explicit, inspectable structure, and more tractable because that structure can in principle be examined, challenged, and corrected in ways that an opaque model cannot.

\section{Grounding the Framework: A High-Stakes Administrative Test Case}

Public administration benefit eligibility decisions are among the most consequential and legally structured applications of AI in contemporary governance. They determine access to housing support, unemployment benefits, disability allowances, and social care for populations that are simultaneously the most dependent on correct decisions and the least equipped to challenge incorrect ones \citep{bovens2002digital,eubanks2018automating}.
 The criteria for eligibility are typically defined by statute, meaning the reasoning structure is not merely desirable but legally mandated: a natural DAG whose nodes correspond to legally defined conditions and whose edges encode the inferential relationships between them. When an automated system makes or informs such a decision, the requirement for auditability and contestability is not a design preference but a legal obligation \citep{citron2014scored, wachter2017counterfactual}.

\begin{figure*}
\centering
\begin{tikzpicture}[font=\sffamily\small, >=Latex,
  node/.style={draw=teal!50, fill=teal!8, rounded corners=4pt,
               minimum width=2.6cm, minimum height=0.65cm,
               align=center}]

\node[draw=teal!70, fill=teal!18, rounded corners=4pt,
      minimum width=2.6cm, minimum height=0.65cm,
      align=center, line width=0.8pt, font=\sffamily\small\bfseries]
     (eli) at (0, 0) {EligibilityOutcome};

\node[node] (emp) at (0,  2) {EmploymentStatus};
\node[node] (inc) at (4, 0)  {IncomeReceived};
\node[node] (res) at (0, -2) {ResidencyStatus};
\node[node] (con) at (-4, 0) {ContributionRecord};

\draw[->, teal!60, line width=0.8pt] (emp) -- (eli);
\draw[->, teal!60, line width=0.8pt] (inc) -- (eli);
\draw[->, teal!60, line width=0.8pt] (res) -- (eli);
\draw[->, teal!60, line width=0.8pt] (con) -- (eli);

\draw[->, teal!60, line width=0.8pt] (emp) -- (inc);

\end{tikzpicture}
\caption{A minimal BN for benefit eligibility
         reasoning. Four statutorily defined condition nodes feed
         into the central eligibility outcome. The edge from
         EmploymentStatus to IncomeReceived reflects a genuine
         conditional dependency: employment status is informative
         about income patterns, and an inconsistency between
         the two signals triggers a targeted re-query.}
\label{fig:eligibility}
\end{figure*}
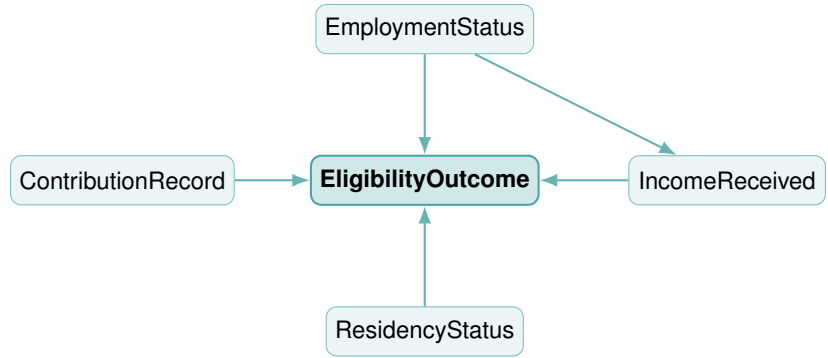

This setting is an ideal test case for the Glassbox Framework precisely because the gap between what current systems provide and what accountability requires is most visible here. An LLM processing an applicant's file can extract relevant signals and generate a plausible eligibility assessment, but it cannot guarantee that the reasoning behind that assessment corresponds to the statutory criteria, remains stable across structurally similar cases, or can be contested at the level of specific legal conditions. A Glassbox system, by contrast, routes the LLM's extractions through a BN encoding the statutory eligibility structure, producing an inference trace that identifies which conditions are satisfied, with what probability, and under what assumptions. Figure~\ref{fig:eligibility} illustrates a minimal BN for this domain.

Table~\ref{tab:trace} illustrates a concrete execution trace for a hypothetical applicant file. The LLM parsing layer extracts candidate variable assignments from the document. The BN evaluates these assignments against the encoded statutory structure and computes posterior probabilities. In this scenario, the LLM extracts a high-confidence signal that the applicant is unemployed, but simultaneously extracts a low-confidence signal suggesting regular income. Given the conditional dependency between EmploymentStatus and IncomeReceived encoded in the DAG, these assignments are structurally incoherent: the BN detects a joint probability anomaly and raises an inconsistency flag. The LLM is re-queried with a targeted prompt focused specifically on the income evidence (e.g. bank statements, employer references, freelance contracts) rather than the full document. On the second pass, the income signal resolves to a clearer state, the inconsistency is cleared, and the BN proceeds to inference. The final output is not a bare eligibility decision but an auditable trace: a structured account of which statutory conditions were assessed, what evidence supported each assessment, and what probability was assigned to the outcome.

\begin{table*}
\centering
\caption{Glassbox execution trace for a benefit eligibility application. The inconsistency flag on IncomeReceived triggers a targeted re-query before inference proceeds. Posterior probabilities are computed by the BN after the second parsing pass.}
\label{tab:trace}
\scalebox{0.82}{
\begin{tabular}{lccc}
\hline
\textbf{BN Node} & \textbf{LLM Extraction} 
  & \textbf{BN Assessment} & \textbf{Posterior} \\
\hline
EmploymentStatus   & Unemployed (high conf.) 
  & Coherent      & $P = 0.91$ \\
IncomeReceived     & Unclear (low conf.)     
  & \textbf{Flag} & \textit{re-queried} \\
ResidencyStatus    & Confirmed resident      
  & Coherent      & $P = 0.97$ \\
ContributionRecord & 18 months confirmed     
  & Coherent      & $P = 0.88$ \\
\hline
IncomeReceived     & Not received (resolved) 
  & Coherent      & $P = 0.83$ \\
\hline
EligibilityOutcome & ---                     
  & Inferred      & $P = 0.87$ \\
\hline
\end{tabular}}
\end{table*}

A standard LLM pipeline processing the same file would produce an eligibility assessment accompanied at best by a post-hoc summary of which document sections were attended to: an account with no formal relationship to the statutory criteria, unstable across similar applications, and impossible to contest at the level of specific legal conditions \citep{citron2014scored}. This trace is not an explanation of what an LLM decided. It is a record of what the BN inferred, under what statutory assumptions, from what evidence. An applicant who disagrees with the outcome can identify a specific node (e.g. ContributionRecord) and challenge the evidence used to instantiate it. An auditor can examine the full trace to verify that the statutory criteria were correctly encoded and correctly applied. A governance body can inspect the DAG to determine whether the eligibility conditions reflect current legislation or require updating following a statutory amendment. These forms of institutional engagement are not possible with post-hoc explanation systems, where the reasoning is reconstructed after the fact by a separate mechanism with no formal relationship to the decision that was made \citep{citron2014scored}.
The governance implication is equally clear. The DAG in Figure~\ref{fig:eligibility} is not merely a technical design choice. The decision to include ContributionRecord as a node, or to encode a conditional dependency between EmploymentStatus and IncomeReceived, is a statutory and institutional decision about what counts as eligibility and how evidence should be weighted. Who makes that decision, under what legislative authority, and through what process of public accountability, is the governance question the Glassbox Framework makes explicit, and that post-hoc explanation systems obscure entirely.

\section{What Glassbox AI Requires}

The Glassbox Framework is not a finished system. It is a research direction whose realisation depends on solving a set of foundational problems that are currently open. We identify five such problems here, each constituting a genuine scientific challenge rather than an engineering task. Together they define the programme that the framework makes necessary.

\subsection{Semantic Grounding} The mapping from natural language to probabilistic variable space requires formal methods that go well beyond lexical matching or embedding-based retrieval. A principled grounding must account for context dependence (the same expression meaning different things in different documents), polysemy at the variable level, negation, hedging, and the systematic failure of language to carve probability space at its joints. The difficulty is not merely practical. It is representational: natural language is designed to communicate in context, while BN variables are designed to abstract away from context. These two purposes are in tension, and no existing framework addresses all three dimensions of the problem simultaneously: formal semantics, probabilistic reasoning, and computational linguistics. Progress here would require new theory of the relationship between linguistic meaning and probabilistic structure, building on but going well beyond current work on grounding language models in structured knowledge \citep{pearl2009causality, koller2009probabilistic}. The thought experiment developed in the previous section illustrates what failure at this interface looks like in practice: a normatively central concept rendered statistically invisible not because the BN was wrong, but because the language pipeline could not reliably identify it.

\subsection{Structure Learning in Normative Domains} In standard structure learning, a DAG is inferred from data subject to statistical criteria: conditional independence tests, score-based optimisation, or constraint-based algorithms \citep{koller2009probabilistic}. In normative domains, this is insufficient on its own. The DAG must also be defensible relative to legal doctrine, institutional policy, or ethical principle, and these requirements are not derivable from data. A model that learns the statistical regularities of past administrative decisions is not the same as a model that encodes the normative principles those decisions were supposed to apply. The two may be extensionally similar in most cases and deeply divergent in the cases that matter most: edge cases, novel circumstances, and situations where the law requires a judgment that past practice got wrong. How do you learn or elicit a structure that is simultaneously data-consistent and normatively legitimate? How do you certify that a proposed DAG faithfully encodes the domain it is meant to represent, and detect when it no longer does as the domain evolves? These questions connect to causal discovery, expert elicitation, and model criticism \citep{fenton2018risk}, but the normative constraint has not been formalised in any of them. Addressing it requires new theory at the boundary of graphical modelling and institutional design, and new evaluation frameworks that go beyond predictive accuracy.

\subsection{Probabilistic Interface} The central mathematical challenge of the Glassbox Framework is constructing a principled mapping between the distributional representations of LLMs and the discrete structured representations of BNs. An LLM assigns probability mass over token sequences; a BN assigns probability mass over variable assignments. These are not two implementations of the same idea: they are incompatible uncertainty regimes. Translating between them in a way that preserves coherence is not solved by existing calibration or uncertainty quantification methods \citep{doshi2017towards, ballester2022computing}. The difficulty goes deeper than calibration in the standard sense. Even if an LLM's confidence scores were perfectly calibrated with respect to empirical frequencies, this would not provide the conditional probability tables a BN requires, because the BN needs probabilities conditioned on specific variable assignments, not scores over a vocabulary. Bridging these regimes requires a formal theory of the interface: what mathematical conditions must a mapping satisfy to allow LLM-derived signals to enter the BN as valid evidence without distorting inference? Under what conditions does such a mapping exist? How is it estimated? These are open questions whose resolution constitutes the mathematical core of the agenda outlined in this paper.

\subsection{Human-in-the-Loop Governance}
The governance layer of the Glassbox Framework requires formal protocols that do not currently exist. How is the DAG specified when domain experts disagree? How are non-technical institutional actors (e.g. judges, administrators, patients) equipped to audit a BN structure they did not build? How can affected parties contest specific nodes or conditional dependencies, and through what procedural mechanism is that challenge resolved? How is the model updated when legal standards change or new evidence types emerge, and who authorises that update? These questions must be answered at two levels simultaneously: procedurally, in ways that satisfy the institutional legitimacy requirements of the deployment context \citep{kroll2017accountable}, and technically, in ways that preserve the coherence of the probabilistic reasoning structure \citep{raji2020closing}. High-level ethical principles are insufficient here \citep{mittelstadt2019principles}. What is needed is a formal institutional architecture for probabilistic model governance that does not currently exist. Designing it requires collaboration across law, institutional theory, and probabilistic modelling at a level of precision that the existing literature has not yet achieved.

\subsection{Scalability and Generalisation} The framework as described is instantiated through domain-specific BNs. A BN for homicide liability, a BN for benefit eligibility, and a BN for medical triage have different variable sets, different causal structures, different evidentiary logics, and different governance requirements. Moving between domains is not a matter of swapping variable sets and retraining. It raises the question of whether and how knowledge from one instantiation can inform another: whether there exist meta-level structures, modular components, or transfer mechanisms that generalise across deployments \citep{neil2000building}. Without progress here, the Glassbox Framework risks becoming a collection of domain-specific engineering projects. The conditions under which a BN encoding from one domain can be adapted to another (preserving the governance properties and interpretability guarantees that justify the framework in the first place) are neither obvious nor trivial, and their investigation would constitute a substantive contribution to both probabilistic modelling and AI governance.

\section{Conclusions}
The dominant response to the opacity of AI systems in high-stakes settings has been to explain outputs after the fact. This paper has argued that this response is structurally inadequate: post-hoc explanation cannot provide the stability, contestability, or formal accountability that institutional deployment requires, and the emergence of LLMs as infrastructural components has made the alternative of simply using interpretable models unavailable in many of the domains where accountability matters most.

The Glassbox Framework proposes a different path. By positioning BNs as ante-hoc reasoning layers that mediate between unstructured language and auditable institutional outputs, the framework shifts accountability from an afterthought to a design principle. The reasoning structure is not reconstructed after inference. It is specified before inference begins, encoded in an inspectable probabilistic model whose assumptions can be examined, challenged, and corrected. The governance layer makes explicit who is responsible for that structure. The accountability layer makes the full inference trace available for institutional review. The result is a system whose outputs are not merely explained but genuinely contestable.

This paper does not claim to have built this system. What it claims is more precise and, we believe, more valuable: that the framework is the right architecture for accountable generative AI, that the interface between language and probability is the central scientific object whose properties must be understood, and that the five foundational challenges identified in the preceding section constitute a coherent and non-trivial scientific agenda. Each challenge is hard in a specific and characterisable way. That precision is itself a contribution.

As LLMs become the reasoning substrate of consequential institutions, the question of whether their outputs can be genuinely accounted for becomes urgent. Post-hoc explanation cannot answer it. Probabilistic mediation, if the foundational problems are solved, can.

\section{Acknowledgments}
This  paper  was  partially  funded  by  PID2023-153222OB-I00  granted  by  MCIU  /  AEI  /  10.13039/501100011033  /FEDER, UE.

\bibliography{aaai2026}


\end{document}